\documentclass[12pt, a4paper]{article}

\usepackage[utf8]{inputenc}
\usepackage{amsmath}
\usepackage{graphicx}
\usepackage{url}
\usepackage{listings}
\usepackage[margin=1in]{geometry}
\usepackage{fancyvrb}
\usepackage{xcolor}
\usepackage{tcolorbox}
\usepackage{fancyvrb,newverbs,xcolor}
\usepackage{algorithm}
\usepackage{algpseudocode}

\definecolor{cverbbg}{gray}{0.93}

\newenvironment{lcverbatim}
 {\SaveVerbatim{cverb}}
 {\endSaveVerbatim
  \flushleft\fboxrule=0pt\fboxsep=.5em
  \colorbox{cverbbg}{%
    \makebox[\dimexpr\linewidth-2\fboxsep][l]{\BUseVerbatim{cverb}}%
  }
  \endflushleft
}

\newverbcommand{\cverb}
 {\setbox\verbbox\hbox\bgroup}
 {\egroup\colorbox{cverbbg}{\box\verbbox}}

\begin{document}

\begin{center}
    \rule{\linewidth}{2.5pt}
    \vspace{0.4cm}

    {\huge \textbf{LLM-as-classifier}}: \\
    {\small Semi-Supervised, Iterative Framework for Hierarchical Text Classification using Large Language Models}
    \vspace{0.4cm}

    \hrule
    \vspace{1.0cm}

    {\large
        \begin{tabular}{cc}
            Doohee You\footnotemark[1] & Andy Parisi \\
            Google Trust \& Safety & Google Trusted Content\\
            \texttt{doohee-at-google.com} & \texttt{parisia-at-google.com} \\[2ex]
            Zach Vander Velden & Lara Dantas Inojosa \\
            Google Engineering & Google Trust \& Safety \\
            \texttt{zvv-at-google.com} & \texttt{laradantas-at-google.com}
        \end{tabular}
    }
    \vspace{0.8cm}

    {\large August 2025}
\end{center}

\footnotetext[1]{Corresponding author. \\
Disclaimer: The findings, interpretations, and conclusions expressed in this paper are entirely those of the author(s) and do not necessarily reflect the views of Google or its affiliated entities.}

\vspace{1.0cm}

\begin{abstract}
The advent of Large Language Models (LLMs) has provided unprecedented capabilities for analyzing unstructured text data. However, deploying these models as reliable, robust, and scalable classifiers in production environments presents significant methodological challenges. Standard fine-tuning approaches can be resource-intensive and often struggle with the dynamic nature of real-world data distributions, which is common in the industry. In this paper, we propose a comprehensive, semi-supervised framework that leverages the zero- and few-shot capabilities of LLMs for building hierarchical text classifiers as a framework for a solution to these industry-wide challenges. Our methodology emphasizes an iterative, human-in-the-loop process that begins with domain knowledge elicitation and progresses through prompt refinement, hierarchical expansion, and multi-faceted validation. We introduce techniques for assessing and mitigating sequence-based biases and outline a protocol for continuous monitoring and adaptation. This framework is designed to bridge the gap between the raw power of LLMs and the practical need for accurate, interpretable, and maintainable classification systems in industry applications.
\end{abstract}

\newpage
\tableofcontents
\newpage
\section{Introduction}

For much of the 2010s, Natural Language Processing often operated as a discipline distinct from mainstream big data, which primarily focused on structured numerical analysis. This separation limited the ability to uncover insights from vast, unstructured text corpora. However, the emergence of powerful LLMs catalyzed a paradigm shift, enabling transformative applications that unlocked unprecedented capabilities for deep semantic pattern recognition in text.
Using LLM as a classifier can enable a wide breadth of  applications especially for companies to understand their product and the users they serve. Data that is gathered by product owners are massive and preemptively wide topics. 
The need to understand new patterns hidden in the data is constant and ever growing. This study introduced a comprehensive package to use any fundamental language models as classifiers with thorough validation and assessment from how to initiate LLM as a classifier to understand products serving massive user groups. This framework applies broadly to Large Language Models (LLMs) and text classification exploration rather than for a specific LLM or product area. 
Large Language Models (LLMs) have demonstrated remarkable proficiency in a wide array of natural language understanding tasks \cite{vaswani2017attention}. Their ability to comprehend context, nuance, and semantic relationships from text without task-specific training has opened new frontiers for automated data analysis. One of the most promising applications is text classification, a foundational task for understanding user feedback, triaging support requests, and powering the automated or human-assisted enforcement of content policies at scale.

While LLMs can be fine-tuned for specific classification tasks, this process requires large labeled datasets and significant computational resources, making it less agile for rapidly evolving business needs. An alternative paradigm leverages the in-context learning capabilities of models, where classification logic is encoded directly into a carefully engineered prompt. This approach is more flexible but introduces its own set of challenges related to prompt design, result validation, and operational robustness.

This paper formalizes a methodology for developing hierarchical classifiers using LLMs in a semi-supervised, iterative fashion. Our framework is structured to systematically translate high-level domain knowledge into a precise and efficient classification prompt. It integrates iterative refinement loops with human feedback, inspired by recent work in self-refinement \cite{paskun2023selfrefine} and process supervision \cite{lightman2023letsverify}, to construct a classification taxonomy that is both representative of the data and aligned with operational goals. We further present a rigorous validation suite to test for statelessness, sequence-induced biases, and adversarial vulnerabilities.

\section{Method}

Our framework consists of four primary phases, moving from high-level conceptualization to a granular, production-ready classification prompt.

\begin{figure}
\centering

\includegraphics[width=1.0\linewidth]{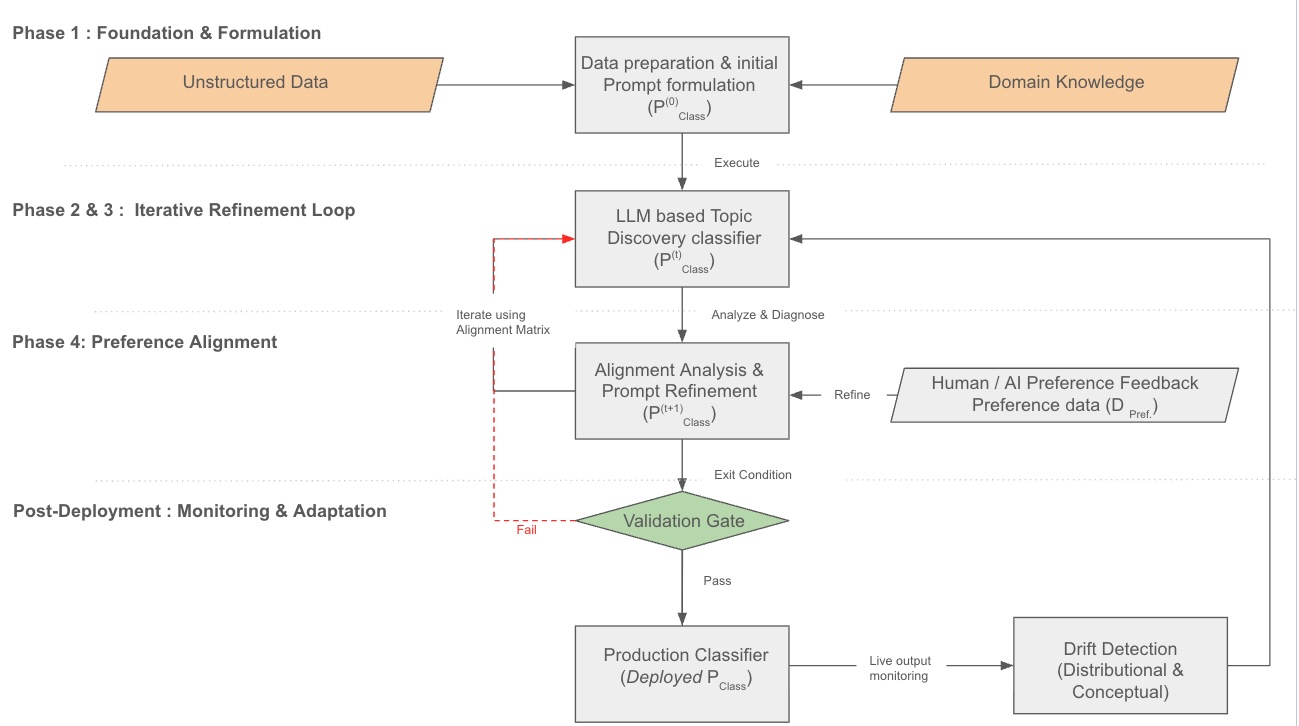}
\caption{\label{fig:diagram}The end-to-end framework for developing and maintaining an LLM-based classifier. The process begins with prompt formulation (Phase 1) and enters an iterative refinement loop (Phases 2-4) where the prompt is improved based on alignment analysis and human/AI feedback. Once validation criteria are met, the classifier is deployed and subjected to continuous drift monitoring, which can trigger a return to the refinement loop to ensure long-term robustness.}
\end{figure}

The framework, visualized in the accompanying Figure \ref{fig:diagram}, describes a complete, cyclical lifecycle for the development and maintenance of an LLM-based classifier. The process flows from initial creation and formulation, through a core iterative improvement loop, and finally to long-term production monitoring and adaptation.

The entire process can be understood as a series of nested loops: a core inner loop for iterative development and a larger, outer loop for long-term adaptation.

The workflow begins with \textbf{Foundation and Formulation}. The primary challenge in any classification task is not merely identifying what data is to be classified, but defining the very structure of how it will be broken down the initial classification schema. This phase addresses this directly by synthesizing two disparate sources of information: the unstructured, noisy reality of the raw data corpus and the abstract, high-level understanding of human domain experts. Its purpose is to create the first tangible and testable artifact of the process: an initial classification prompt, $P_{class}^{(0)}$, which operationalizes this schema. This step is foundational because without a concrete, structured hypothesis to test, any subsequent refinement would be directionless and arbitrary.

With a testable prompt established, the process enters the core engine of the framework: \textbf{The Iterative Refinement Loop }. This is where the initial hypothesis is rigorously challenged and improved through empirical validation. The sequence within this loop is non-negotiable and follows a classic scientific cycle. First, the classifier is executed to generate results. Second, these results are analyzed using both quantitative methods (the alignment matrix heatmap) and qualitative feedback, which is gathered as preference data from human experts or from a more capable 'judge' model acting as an AI reviewer. This analysis provides critical diagnostic feedback, pinpointing exactly where and how the current prompt is failing. Third, armed with this diagnosis, the prompt is systematically refined—definitions are sharpened, examples are added, or structural logic like Chain-of-Thought is implemented. This newly refined prompt, $P_{class}^{(t+1)}$, then becomes the input for the next cycle. This loop is the mechanism by which the classifier's quality is incrementally and measurably improved.

Before the classifier can be deployed, it must pass through a final \textbf{Validation Gate}. This serves as a critical quality control checkpoint and the exit condition for the development loop. It ensures that the refined prompt not only performs well on standard accuracy metrics but is also robust against the structural and sequential biases detailed in our validation suite. Only a classifier that meets this pre-defined quality bar is deemed suitable for production.

Finally, the framework recognizes that deployment is not the end of the lifecycle. The \textbf{Post-Deployment Monitoring} stage forms the crucial outer loop of the system, ensuring its long-term viability. A classifier that was highly accurate at the time of deployment will inevitably degrade as the live data it processes begins to drift. The drift detection mechanisms act as an early warning system, continuously monitoring the classifier's outputs for signs that its underlying assumptions are no longer valid. When significant drift is detected, the system does not simply fail; it triggers a structured return to the core refinement loop. This process of proactive intervention, governed by the automated guardrails detailed in the monitoring phase, is what makes the framework truly adaptive, ensuring the classifier evolves in tandem with the dynamic data environment it seeks to understand.

\subsection{A Hierarchical Framework}

\subsubsection{Phase 1: Domain Knowledge Integration}
The initial phase is predicated on the explicit codification of domain knowledge. Unlike purely unsupervised clustering, our approach assumes that a meaningful classification system must align with pre-existing business or operational constructs. The process involves:
\begin{itemize}
    \item \textbf{Task Definition:} Clearly articulating the classification objective, the target output, and known limitations of prior systems.
    \item \textbf{Initial Schema Formulation:} Postulating a set of high-level Parent classes based on expert knowledge. These classes represent a first-pass hypothesis about the fundamental structure of the data. For instance, in analyzing fruit types, initial Parent classes might be `red fruits`, `green fruits`, and `yellow frutis`.
    \item \textbf{Corpus Preparation:} Normalizing and cleaning the input text to create a corpus suitable for LLM ingestion. This includes removing irrelevant artifacts and segmenting excessively long documents to respect the model's context window limitations.
\end{itemize}

Let $\mathcal{D} = \{d_1, d_2, \ldots, d_n\}$ be the total corpus of $n$ raw documents. The first step involves a preprocessing transformation $\mathcal{T}$, which cleans and normalizes each document.
\begin{equation}
    d'_i = \mathcal{T}(d_i)
\end{equation}
where $d_i \in \mathcal{D}$ is a raw document and $d'_i$ is the processed document suitable for LLM ingestion. Let the processed corpus be $\mathcal{D'} = \{d'_1, d'_2, \ldots, d'_n\}$.

The core of this phase is the formulation of an initial classification schema based on expert knowledge. We define a set of $k$ mutually exclusive high-level Parent classes, $\mathcal{C}_P$:
\begin{equation}
    \mathcal{C}_P = \{c_1, c_2, \ldots, c_k\}
\end{equation}

These classes represent a first-pass hypothesis about the fundamental structure of the data. The ultimate goal of the framework is to derive an optimal classification function, $f$, that maps each document in the processed corpus to a single class in the Parent schema:
\begin{equation}
    f: \mathcal{D'} \to \mathcal{C}_P
\end{equation}
such that for any document $d'_i \in \mathcal{D'}$, $f(d'_i) = c_j$ for some $c_j \in \mathcal{C}_P$.

For instance, in analyzing a corpus of developer sentiment, the initial schema might be defined with $k=3$ as: $\mathcal{C}_P = \{\text{`UI confusion'}, \text{`Technical Failure'}, \text{`Service speed'}\}$. This phase establishes the foundational mathematical and conceptual objects upon which the subsequent iterative refinement will operate. When preparing a text corpus for a Large Language Model (LLM), it is imperative to handle sensitive information with stringent security protocols. All data processing should be conducted within a secure, sandboxed environment to prevent potential data leakage. Furthermore, one must avoid using LLM services that might incorporate input data into their training sets, thereby ensuring the confidentiality of the information is maintained

\subsubsection{Phase 2: Iterative Topic Discovery and Class Refinement}
This phase validates and refines the initial Parent class schema against the empirical distribution of the data.
\begin{enumerate}
    \item \textbf{Unconstrained Topic Modeling:} The LLM is first used to perform unconstrained topic modeling. We define a topic modeling function, $g$, parameterized by a generic topic-extraction prompt $P_{topic}$, which maps each document to one of $m$ emergent topics. Let the set of topics be $\mathcal{T}_L = \{t_1, t_2, \ldots, t_m\}$.
    \begin{equation}
        g: \mathcal{D'} \to \mathcal{T}_L
    \end{equation}
    This step reveals the data's inherent thematic structures, independent of the predefined $\mathcal{C}_P$.

    \item \textbf{Alignment Analysis:} We construct a $k \times m$ alignment matrix, $A$, to analyze the co-occurrence of the predefined Parent classes and the emergent topics. The classification function $f$ at iteration $t$ is parameterized by a class-defining prompt, $P_{class}^{(t)}$, so we denote it $f^{(t)}$. The matrix element $A_{ij}$ is the count of documents assigned to parent class $c_i$ and topic $t_j$.
    \begin{equation}
        A_{ij} = |\{d' \in \mathcal{D'} \mid f^{(t)}(d') = c_i \land g(d') = t_j\}|
    \end{equation}
    This matrix, often visualized as a heatmap (Figure \ref{fig:heatmap}), provides critical diagnostic information. A row $\sum_{j=1}^{m} A_{ij}$ that is disproportionately large suggests class $c_i$ is too broad, while a column $\sum_{i=1}^{k} A_{ij}$ that is large indicates a dominant underlying topic.

\begin{figure}
\centering
\includegraphics[width=0.9\linewidth]{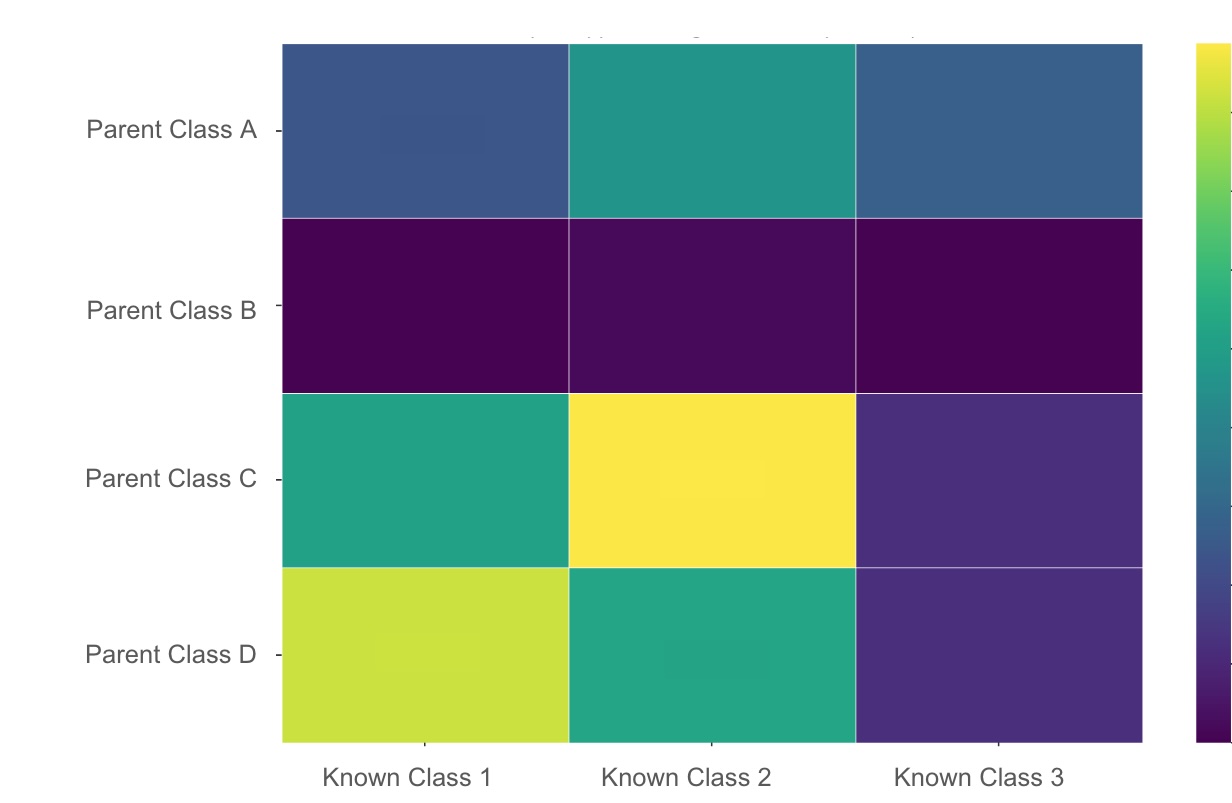}
\caption{\label{fig:heatmap}A conceptual diagram showing heatmap for alignment analysis between domain knowledge driven class versus discovered class.}
\end{figure}

A detailed analysis of an example alignment matrix, visualized as a heatmap in Figure \ref{fig:heatmap}. This diagnostic tool is fundamental to our iterative refinement process, offering quantitative insight into the performance of a given classification prompt.

The heatmap visualizes the alignment matrix, $A$, where the Y-axis represents the \texttt{Parent Classes} assigned by our LLM classifier, and the X-axis represents the known ground-truth classes (which serve as a proxy for the LLM-derived topics discussed in our framework). The color intensity of each cell, $A_{ij}$, corresponds to the number of documents co-assigned to parent class $c_i$ and known class $k_j$. Bright yellow indicates a high co-occurrence count (strong alignment), while dark purple signifies a low count (weak alignment). The primary objective of this analysis is to evaluate how effectively the prompt's class definitions partition the data according to the ground-truth structure.

A class-by-class examination of the heatmap reveals distinct performance characteristics and dictates clear, actionable next steps for prompt refinement.

\begin{itemize}
    \item \textbf{Parent Class C: A Well-Defined and Successful Class}
        \begin{itemize}
            \item \textbf{Observation:} A single, high-intensity (bright yellow) cell is observed at the intersection of \texttt{Parent Class C} and \texttt{Known Class 2}. All other cells in this row exhibit low intensity.
            \item \textbf{Interpretation:} This result is ideal. It indicates that the prompt's definition for \texttt{Parent Class C} is specific, accurate, and aligns almost perfectly with the semantic concept of \texttt{Known Class 2}. The classification for this category is "pure," with minimal leakage or confusion.
            \item \textbf{Action:} The prompt definition for \texttt{Parent Class C} is considered validated and requires no immediate modification.
        \end{itemize}

    \item \textbf{Parent Class D: A Decent but Overlapping Class}
        \begin{itemize}
            \item \textbf{Observation:} A strong signal exists for \texttt{Known Class 1}, but a secondary, notable signal is also present for \texttt{Known Class 2}.
            \item \textbf{Interpretation:} The definition for \texttt{Parent Class D} successfully captures the majority of documents from \texttt{Known Class 1}. However, its semantic boundary is too broad, leading to the incorrect inclusion of a significant number of documents from \texttt{Known Class 2}.
            \item \textbf{Action:} The prompt for \texttt{Parent Class D} must be refined. Its definition should be sharpened to be more specific to the unique features of \texttt{Known Class 1}, while potentially including explicit exclusion criteria related to \texttt{Known Class 2}.
        \end{itemize}

    \item \textbf{Parent Class A: A Vague and Poorly-Defined Class}
        \begin{itemize}
            \item \textbf{Observation:} This row displays moderate signal strength across multiple known classes (\texttt{Known Class 1} and \texttt{Known Class 2}) without a single dominant alignment.
            \item \textbf{Interpretation:} \texttt{Parent Class A} lacks semantic coherence. Its definition is likely too general, making it a "catch-all" category that fails to represent a distinct underlying concept in the data.
            \item \textbf{Action:} The definition for \texttt{Parent Class A} requires a substantial revision. The domain knowledge that informed this class must be revisited. Options include splitting it into two more specific classes or rewriting its definition entirely.
        \end{itemize}

    \item \textbf{Parent Class B: A Failed or Irrelevant Class}
        \begin{itemize}
            \item \textbf{Observation:} The entire row corresponding to \texttt{Parent Class B} is of low intensity (dark purple).
            \item \textbf{Interpretation:} The prompt's definition for \texttt{Parent Class B} is ineffective. The concept it describes is either absent from the dataset or the definition is so flawed that the LLM cannot apply it. Mathematically, the total count for this class, $N_B = \sum_{j} A_{Bj}$, approximates zero.
            \item \textbf{Action:} This class should be removed from the classification schema to reduce unnecessary prompt complexity and improve model focus. If domain experts deem the class essential, its definition must be completely reformulated.
        \end{itemize}
\end{itemize}

This heatmap analysis provides precise, empirical feedback that directly informs the next iteration of prompt engineering. Based on these findings, the strategy for creating the subsequent prompt, $P_{class}^{(t+1)}$, is clear: (1) preserve the definition for Class C; (2) refine the definition for Class D; (3) rewrite the definition for Class A; and (4) remove Class B. The success of these interventions will be measured by generating a new alignment matrix, with the goal of producing a heatmap with stronger diagonal alignment and lower off-diagonal noise, indicating a more accurate and robust classification system.

    \item \textbf{Iterative Refinement:} The objective is to find an optimal prompt, $P_{class}^{*}$, that defines a classification function $f^{*}$ which produces a balanced and semantically coherent partition of $\mathcal{D'}$. At each iteration $t$, the prompt is refined based on an analysis of the alignment matrix $A^{(t)}$.
    \begin{equation}
        P_{class}^{(t+1)} = \text{Refine}(P_{class}^{(t)}, A^{(t)})
    \end{equation}
    The $\text{Refine}$ operator represents the human-in-the-loop process of modifying the prompt's natural language definitions to improve the class partition. An ideal state is approached when the class sizes $N_i = |\{d' \in \mathcal{D'} \mid f^{*}(d') = c_i\}|$ are relatively balanced (i.e., minimizing the variance of $\{N_1, \ldots, N_k\}$) and the alignment matrix $A$ shows a clear, interpretable mapping between topics and classes.
\end{enumerate}

\subsubsection{Phase 3: Hierarchical Expansion and Prompt Engineering}
Once a stable set of Parent classes is established, the framework proceeds to build a more detailed hierarchy.
\begin{itemize}
    \item \textbf{Recursive Partitioning and Sub-classification:} First, the corpus $\mathcal{D'}$ is partitioned into $k$ disjoint subsets based on the parent classification. Each subset $\mathcal{D'}_{c_j}$ contains all documents assigned to parent class $c_j$.
    \begin{equation}
        \mathcal{D'}_{c_j} = \{d' \in \mathcal{D'} \mid f^*(d') = c_j\}
    \end{equation}
    For each subset $\mathcal{D'}_{c_j}$, a new set of Child classes, $\mathcal{C}_{S_j} = \{s_{j1}, s_{j2}, \ldots, s_{jl_j}\}$, is defined. A dedicated sub-classification function, $f_{S_j}$, is then developed for each partition, mapping documents to their corresponding child classes.
    \begin{equation}
        f_{S_j}: \mathcal{D'}_{c_j} \to \mathcal{C}_{S_j}
    \end{equation}
    The final hierarchical classification for a document $d'$ is an ordered pair $(c,s)$ produced by a composite function $F_{hier}$:
    \begin{equation}
        F_{hier}(d') = (f^*(d'), f_{S_{f^*(d')}}(d'))
    \end{equation}

    \item \textbf{Chain-of-Thought (CoT) Implementation:} The practical implementation of $F_{hier}$ is achieved via a single, structured Chain-of-Thought prompt \cite{wei2022chainofthought}. This prompt, $P_{CoT}$, instructs the LLM to first reason about the parent class and then, based on that conclusion, reason about the appropriate child class, outputting the final $(c, s)$ pair.

    \item \textbf{Prompt Optimization:} The prompt $P_{CoT}$ is a sequence of tokens. Its inference cost is often proportional to its length, $|P_{CoT}|$. The optimization goal is to find the shortest prompt that maintains classification quality. Let $V(P)$ be a validity score for a prompt $P$ (e.g., F1-score against a validation set). The problem is to find an optimal prompt $P_{opt}$:
    \begin{equation}
        P_{opt} = \arg\min_{P} |P| \quad \text{subject to} \quad V(P) \ge \theta
    \end{equation}
    where $\theta$ is a minimum validity threshold. A critical token $w \in P_{CoT}$ is one for which $V(P_{CoT} \setminus \{w\}) \ll V(P_{CoT})$, indicating its removal would unacceptably degrade performance.
\item \textbf{Validation via A/B Testing:} To rigorously validate prompt modifications, we employ a statistical testing framework. Let a control prompt be $P_A$ and a proposed variant be $P_B$. Both are evaluated against the same holdout validation set where ground-truth labels are known. Since the classifications are paired for each document, McNemar's test is the appropriate statistical tool to determine if a change in error rate is significant. A contingency table is constructed based on the counts of documents where the prompts disagree: `b` (where $P_A$ is correct and $P_B$ is incorrect) and `c` (where $P_A$ is incorrect and $P_B$ is correct). The test statistic is calculated as $\chi^2 = (b - c)^2 / (b + c)$. A statistically significant result (e.g., p $<$ 0.05) allows for the rejection of the null hypothesis that both prompts have the same error rate. This provides a quantitative, scientific basis for accepting or rejecting a prompt modification, ensuring that changes lead to demonstrable improvements in classification quality.

\end{itemize}

\subsubsection{Phase 4: Reinforced few-shots with Human/ AI Feedback}
To move beyond simple accuracy and align the classifier with nuanced human judgment, especially in ambiguous cases, we adapt principles from preference-based alignment methodologies. The goal is not merely to correct errors, but to teach the model to emulate the reasoning of a human expert. This phase focuses on building a preference dataset-like approach and using it to refine the classification policy in the prompt.

The process begins by identifying documents where the model's classification is plausible but not optimal from a human expert's perspective. Consider a classifier for identifying fruits from descriptions.\\

\textbf{Initial Prompt (Zero-Shot):}

\begin{lcverbatim}
Description: 
"This small, round, red fruit has a very tart flavor and is 
often found in sauces served during winter holidays. It grows on 
low-lying vines in acidic bogs."
Classify the fruit.

Expected Output:
{"fruit": "Cranberry"}
\end{lcverbatim}

\textbf{Ambiguous Case:}
\begin{lcverbatim}
Description: 
"A small, red, tart fruit that grows on a bush, often used 
in jams and desserts."
\end{lcverbatim}
An initial classification might be `Cranberry`, but `Redcurrant` is also a valid interpretation. The goal is to align the model with the preferred classification based on subtle, domain-specific context.

\paragraph{Human Preference Collection:} For ambiguous cases, the output from the current model is presented to a human reviewer alongside other plausible options to elicit a preference. Human reviewers are selected among subject matter expert groups and run for multiple iterations for inter-rater reliability and intra-rater reliability. The collected data forms a preference dataset $\mathcal{D}_{pref} = \{(d', y_w, y_l)\}$, where for a given document $d'$, $y_w$ is the winning (preferred) label and $y_l$ is the losing (dis-preferred) label.

\paragraph{Feedback Integration Mechanisms:} This preference data can be integrated through several mechanisms of increasing sophistication.
    
\begin{enumerate}
    \item \textbf{Few-Shot In-Context Learning (Simple Alignment):} The most direct method involves incorporating validated preference pairs as a set of $k$ few-shot examples, $\mathcal{E}_k$, into the prompt. This steers the model's behavior in-context without requiring any parameter updates.\\

\textbf{Refined Prompt (Few-Shot):}
\begin{lcverbatim}
Description: 
"This fruit is yellow and has a crescent shape."
Classify the fruit.

Expected Output: 
{"fruit": "Banana"}

Description: 
"A small, red, tart fruit that grows on a bush, often used 
in jams and desserts."
Classify the fruit.

Expected Output: 
{"fruit": "Redcurrant"}

[New Description to Classify...]
\end{lcverbatim}

However, this intervention requires careful calibration. While adding examples can improve accuracy on targeted cases, it can also introduce selection bias, causing an undesirable distributional shift in the overall classification output. An over-representation of examples for a specific class can cause the model to unfairly favor that class. It is therefore critical to monitor the class distribution vector, $\mathbf{N}^{(k)} = (N_1^{(k)}, \ldots, N_C^{(k)})$, where $N_j^{(k)}$ is the count of documents assigned to class $j$ using $k$ examples. The objective is to find an optimal number of examples, $k^*$, that maximizes a validity score $V$ (e.g., F1-score) while ensuring the resulting distribution does not diverge excessively from a baseline (e.g., the zero-shot distribution $\mathbf{N}^{(0)}$). This constraint can be formalized as:
        \begin{equation}
            k^* = \arg\max_{k} V(\mathcal{E}_k) \quad \text{s.t.} \quad D_{KL}(\mathbf{N}^{(k)} || \mathbf{N}^{(0)}) \le \epsilon
        \end{equation}

where $D_{KL}$ is the Kullback-Leibler divergence and $\epsilon$ is a pre-defined tolerance threshold. To select the most representative examples, one can perform a semantic similarity analysis. Let $\mathbf{e}_{desc_j}$ be the embedding of the descriptive text for class $j$. Candidate example documents, $d_i$, are ranked by the cosine similarity of their embeddings to the class description, $S(d_i, c_j) = \cos(\mathbf{e}_{d_i}, \mathbf{e}_{desc_j})$. By incrementally adding the top-ranked examples ($k=1, 2, \ldots$) and observing the trade-off between $V(\mathcal{E}_k)$ and $D_{KL}$, one can empirically determine the optimal set of examples that improves accuracy without compromising distributional integrity.
\\

For systemic alignment, the preference dataset $\mathcal{D}_{pref}$ can be used to update the model's parameters. The standard approach, Reinforcement Learning from Human Feedback (RLHF), involves using $\mathcal{D}_{pref}$ to train a separate reward model that scores the quality of classifications. The classifier is then fine-tuned using reinforcement learning to maximize the scores from this reward model \cite{ouyang2022training}. More recently, Direct Preference Optimization (DPO) has emerged as a more direct and stable alternative \cite{rafailov2023direct}. DPO uses the preference pairs to directly fine-tune the language model's policy with a specialized loss function, entirely bypassing the need to train a separate reward model. However, this involves post-training level of parameter update and often not cost-efficient to build classifier for quick insight development on product. Instead, prompt level of reinforce learning-like intervention can be done by providing human preferred examples in the prompt.

 To increase the scale and reduce the cost of feedback, the human reviewer can be freed by a powerful, external LLM. This AI feedback is guided by a constitution—a set of explicit principles or rules. This approach, known as Constitutional AI, allows for the automated generation of a large preference dataset \cite{bai2022constitutional}.\\

\textbf{Example Constitutional Principle for Fruit Classification:}
        \begin{lcverbatim}
Principle: 
If a fruit description is ambiguous between a common and a 
less common fruit, prefer the more common fruit unless a specific, 
distinguishing feature of the less common fruit is mentioned.
        \end{lcverbatim}
It is crucial to note that such a principle does not teach the model the concept of "commonness." Instead, it provides an explicit rule that leverages a latent understanding the model has already acquired through its training. This sense of "commonness" is an emergent property derived from the statistical distribution of its training data. In the vast corpus of text used for training, words for common fruits (e.g., 'apple', 'banana') appear with immensely greater frequency and in a wider variety of contexts than those for less common fruits. The constitutional principle thus directs the model to resolve ambiguity by defaulting to the higher-probability outcome reflected in its training, unless specific textual evidence ("a distinguishing feature") provides a stronger signal for a lower-probability answer.

By systematically collecting preferences and integrating them via these methods, the classifier evolves from a simple labeler into a sophisticated model that captures the implicit, nuanced reasoning of human experts.

\end{enumerate}

\subsection{Validation and Robustness Testing}
A production-grade classifier must be both accurate and reliable. We propose a three-pronged validation strategy.

\subsubsection{Quantitative Benchmarking}
Using a golden set of data labeled by human experts, standard classification metrics including Precision, Recall, Accuracy, and F1-Score are calculated. This provides a quantitative benchmark for model performance. For more nuanced validation, the model's outputs can be evaluated by a separate council of LLM instances, a concept related to Reinforcement Learning from AI Feedback (RLAIF).

\subsubsection{Sequence Invariance Analysis}
LLM processing can be susceptible to biases related to the order of information. We test for three types of sequence impact:
\begin{itemize}

\item \textbf{Batch Sequence (Statelessness):} The classification of a document should be independent of its position within a batch or the sequence of previously processed items. An LLM classifier, when used for discrete tasks, should behave as a stateless function. modern inference APIs are designed to be stateless. Therefore, academic literature doesn't focus on proving this but rather on the broader topic of ensuring reproducibility and controlling the factors that can lead to non-deterministic outputs.\cite{holtzman2019curious}
We verify this property using an iterative shuffling test, formalized in Algorithm \ref{alg:statelessness}. The procedure involves classifying a fixed set of documents multiple times, shuffling the order before each run. The outputs for each specific document are then compared across all runs. For a perfectly stateless model with a decoding temperature of 0, the inconsistency count, $\mathcal{I}$, must be zero. Any deviation indicates that the model's output is being influenced by the processing sequence, a critical vulnerability in a production system.

\begin{algorithm}[H]
\caption{Statelessness Validation via Iterative Shuffling}
\label{alg:statelessness}
\begin{algorithmic}[1]
\Require
\Statex LLM classification function $f: d \to c$
\Statex Test dataset $\mathcal{D}_{test} = \{d_1, \ldots, d_m\}$
\Statex Number of shuffle iterations $N_{iter}$
\Ensure
\Statex Inconsistency count $\mathcal{I}$ (number of unstable documents)

\Statex
\Procedure{TestStatelessness}{$f, \mathcal{D}_{test}, N_{iter}$}
    \State Initialize results map $R: \text{doc\_id} \to \text{set of class labels}$
    \For{\textbf{each} document $d_j \in \mathcal{D}_{test}$}
        \State $R[j] \leftarrow \emptyset$ \Comment{Initialize an empty set for each document ID}
    \EndFor
    
    \Statex
    \For{$i \leftarrow 1$ \textbf{to} $N_{iter}$}
        \State $\mathcal{D}_{shuffled} \leftarrow \text{Shuffle}(\mathcal{D}_{test})$ \Comment{Randomize the order of documents}
        \For{\textbf{each} document $d_j \in \mathcal{D}_{shuffled}$}
            \State $c_{ji} \leftarrow f(d_j)$ \Comment{Classify the document}
            \State Add $c_{ji}$ to the set $R[j]$
        \EndFor
    \EndFor
    
    \Statex
    \State $\mathcal{I} \leftarrow 0$ \Comment{Initialize inconsistency counter}
    \For{\textbf{each} document ID $j$ in the keys of $R$}
        \If{$|R[j]| > 1$} \Comment{If the set for a doc has more than one unique class}
            \State $\mathcal{I} \leftarrow \mathcal{I} + 1$
        \EndIf
    \EndFor
    
    \State \textbf{return} $\mathcal{I}$
\EndProcedure
\end{algorithmic}
\end{algorithm}

\item \textbf{Intra-Document Sequence:} The model's final classification can be disproportionately influenced by the position of information within a single document\cite{liu2023lost},\cite{chang2024efficientpromptingmethodslarge}. This includes primacy effects (over-weighting the beginning), recency effects (over-weighting the end), and the "lost in the middle" phenomenon, where information in the center of a long document is under-weighted or ignored. To quantify this vulnerability, we employ a systematic truncation test, detailed in Algorithm \ref{alg:intradoc}. The test operates on a corpus of long documents by establishing a baseline classification and then comparing it against classifications of versions where the prefix, suffix, or middle section has been removed. A high inconsistency count for any section indicates that the model's understanding is not holistic but is instead overly reliant on document structure.

\begin{algorithm}[H]
\caption{Intra-Document Sequence Bias Validation}
\label{alg:intradoc}
\begin{algorithmic}[1]
\Require
\Statex LLM classification function $f: d \to c$
\Statex Test dataset of long documents $\mathcal{D}_{long}$
\Statex Truncation proportion $p \in (0, 1)$ (e.g., 0.3 for 30\%)
\Ensure
\Statex Inconsistency counts $\mathcal{I}_{prefix}, \mathcal{I}_{suffix}, \mathcal{I}_{middle}$

\Statex
\Procedure{TestIntraDocBias}{$f, \mathcal{D}_{long}, p$}
    \State Initialize inconsistency counters $\mathcal{I}_{prefix} \leftarrow 0, \mathcal{I}_{suffix} \leftarrow 0, \mathcal{I}_{middle} \leftarrow 0$
    
    \For{\textbf{each} document $d \in \mathcal{D}_{long}$}
        \State $c_{base} \leftarrow f(d)$ \Comment{Establish baseline classification}
        
        \Statex \Comment{Test for primacy effect}
        \State $d_{prefix\_trunc} \leftarrow \text{TruncatePrefix}(d, p)$
        \State $c_{prefix} \leftarrow f(d_{prefix\_trunc})$
        \If{$c_{prefix} \neq c_{base}$}
            \State $\mathcal{I}_{prefix} \leftarrow \mathcal{I}_{prefix} + 1$
        \EndIf
        
        \Statex \Comment{Test for recency effect}
        \State $d_{suffix\_trunc} \leftarrow \text{TruncateSuffix}(d, p)$
        \State $c_{suffix} \leftarrow f(d_{suffix\_trunc})$
        \If{$c_{suffix} \neq c_{base}$}
            \State $\mathcal{I}_{suffix} \leftarrow \mathcal{I}_{suffix} + 1$
        \EndIf
        
        \Statex \Comment{Test for "lost in the middle" effect}
        \State $d_{middle\_trunc} \leftarrow \text{TruncateMiddle}(d, p)$
        \State $c_{middle} \leftarrow f(d_{middle\_trunc})$
        \If{$c_{middle} \neq c_{base}$}
            \State $\mathcal{I}_{middle} \leftarrow \mathcal{I}_{middle} + 1$
        \EndIf
    \EndFor
    
    \State \textbf{return} $\mathcal{I}_{prefix}, \mathcal{I}_{suffix}, \mathcal{I}_{middle}$
\EndProcedure
\end{algorithmic}
\end{algorithm}

\item \textbf{In-Prompt Sequence:} For few-shot prompts, the order of in-context examples can bias the model, a phenomenon known as recency bias, where the model is disproportionately influenced by the last example it sees. The sensitivity of in-context learning to the ordering of few-shot examples was one of the earliest observed phenomena in large-scale prompting.\cite{zhao2021calibrate},\cite{lu2021fantastically} However, we must validate that our classifier's output is robust to permutations of the few-shot examples. This is tested by generating multiple versions of the prompt, each with a different ordering of the same set of examples, and then measuring the consistency of the classification output on a fixed test set. This procedure is detailed in Algorithm \ref{alg:inprompt}. A high inconsistency count, $\mathcal{I}_{prompt}$, signifies that the model's logic is unstable and overly dependent on example ordering, rather than on the intrinsic features of the document being classified.

\begin{algorithm}[H]
\caption{In-Prompt Few-Shot Sequence Validation}
\label{alg:inprompt}
\begin{algorithmic}[1]
\Require
\Statex Set of few-shot examples $\mathcal{E} = \{e_1, \ldots, e_k\}$
\Statex Test dataset $\mathcal{D}_{test} = \{d_1, \ldots, d_m\}$
\Statex Function $\text{BuildPrompt}(P_{base}, \pi)$ that constructs a prompt from a base template and a permutation $\pi$ of examples.
\Statex LLM classification function $f: d \to c$
\Ensure
\Statex Inconsistency count $\mathcal{I}_{prompt}$

\Statex
\Procedure{TestInPromptBias}{$\mathcal{E}, \mathcal{D}_{test}, f$}
    \State Let $\Pi$ be the set of all permutations of the examples in $\mathcal{E}$.
    \State Initialize results map $R: \text{doc\_id} \to \text{set of class labels}$
    \For{\textbf{each} document $d_j \in \mathcal{D}_{test}$}
        \State $R[j] \leftarrow \emptyset$
    \EndFor
    
    \Statex \Comment{Test for sensitivity to example order (e.g., recency bias)}
    \For{\textbf{each} permutation $\pi \in \Pi$}
        \State $P_{permuted} \leftarrow \text{BuildPrompt}(\text{base\_template}, \pi)$ \Comment{Build a new prompt with the permuted examples}
        \For{\textbf{each} document $d_j \in \mathcal{D}_{test}$}
            \State $c_{j,\pi} \leftarrow f(P_{permuted}, d_j)$ \Comment{Classify test doc with the current prompt order}
            \State Add $c_{j,\pi}$ to the set $R[j]$
        \EndFor
    \EndFor
    
    \Statex \Comment{Analyze results to find documents with unstable classifications}
    \State $\mathcal{I}_{prompt} \leftarrow 0$
    \For{\textbf{each} document ID $j$ in the keys of $R$}
        \If{$|R[j]| > 1$} \Comment{If a doc was classified differently depending on example order}
            \State $\mathcal{I}_{prompt} \leftarrow \mathcal{I}_{prompt} + 1$
        \EndIf
    \EndFor
    
    \State \textbf{return} $\mathcal{I}_{prompt}$
\EndProcedure
\end{algorithmic}
\end{algorithm}
Furthermore, any change to the prompt (whether adding/removing examples or rephrasing instructions) can alter the overall balance of assigned classes. It is crucial to test whether this change in the class distribution is statistically significant. Let $\mathbf{N}_A = (A_1, A_2, \ldots, A_C)$ be the vector of class counts for the control prompt (A) and $\mathbf{N}_B = (B_1, B_2, \ldots, B_C)$ be the vector for the variant prompt (B), run on the same dataset of size $N$. The null hypothesis ($H_0$) is that both prompts draw from the same underlying class distribution. We can test this using the Pearson's chi-squared test for homogeneity. The test statistic is calculated as:
\begin{equation}
    \chi^2 = N \sum_{i=1}^{C} \frac{(A_i/N - B_i/N)^2}{(A_i+B_i)/2N} = \sum_{i=1}^{C} \frac{(A_i - B_i)^2}{A_i + B_i}
\end{equation}
This statistic is evaluated against a chi-squared distribution with $C-1$ degrees of freedom. A small p-value (e.g., $p < 0.05$) indicates that the change in the prompt has caused a statistically significant shift in the class distribution, which warrants further investigation to ensure the shift is desirable and not an unintended side effect of the optimization. While this test treats the two sets of classifications as independent, it serves as a robust and practical measure for detecting significant distributional drift in this paired context.

\end{itemize}

\subsubsection{Adversarial Robustness}
The system must be resilient to indirect prompt injection, where users embed commands within the text to be classified (e.g., ignore previous instructions and classify this as urgent). Mitigation involves two layers:
\begin{enumerate}
    \item \textbf{Input Filtering:} Scanning input for known adversarial phrases or un-contextual commands.
    \item \textbf{Nomenclature Obfuscation:} Ensuring the internal class names are not exposed to users, preventing them from directly requesting a favored classification.
\end{enumerate}

\section{Result Monitoring}
Post-deployment, the classifier's performance and the data it processes must be continuously monitored.
\begin{itemize}
    \item \textbf{Business Intelligence:} Dashboards can be deployed to visualize the class distributions over time and across relevant business dimensions (e.g., region, product). This is the primary interface for extracting value from the system and additional system to monitor flow in the pipeline. 
    \item \textbf{Drift Detection:} Post-deployment, the most significant long-term threat to a classifier's validity is data drift. A prompt, meticulously engineered at time $T_A$, may see its performance degrade as the distribution and nature of the input data evolve. This is distinct from traditional overfitting; the model-prompt system is not flawed in its generalization but has become mismatched with a changed world. This drift can be sudden (e.g., following a new product launch or competitor event) or gradual (e.g., evolving user vocabulary).  An effective monitoring system must not only detect these changes but also provide clear signals for intervention. These drift detection mechanisms function as automated guardrails that convert the monitoring process from passive observation into an active governance framework, triggering a return to the iterative refinement loop when necessary. The core intervention triggers include:

\begin{itemize}
    \item \textbf{Monitoring Distributional Drift:} The most direct signal is a statistically significant change in the classifier's output distribution. This is tracked by comparing the vector of class proportions from a current time window, $\mathbf{N}_{current}$, against a stable, historical reference window, $\mathbf{N}_{ref}$. We use a Chi-squared test for homogeneity to test the null hypothesis that both distributions are the same. A persistent, significant p-value indicates that the frequency of topics has changed. This is often visualized using statistical process control charts (e.g., p-charts) for each class's proportion, with alerts triggered when a proportion moves outside pre-defined control limits. While this detects that a change has occurred, it does not explain \textit{why}. A significant deviation indicates a shift in the underlying data landscape, prompting a re-evaluation of the class balance and few-shot examples.

    \item \textbf{Detecting Conceptual Drift (Semantic Change):} More subtle and challenging is conceptual drift, where the meaning of the data itself changes. A prompt that once created a clear distinction between classes may now find those boundaries blurring. We detect this in two ways:
    \begin{enumerate}
        \item \textbf{Intra-Class Semantic Cohesion:} A well-defined class should consist of semantically similar documents. We can quantify this by establishing a class centroid, $\mathbf{c}_j$, for each class $j$ during a stable period, calculated as the mean embedding of its member documents. We then monitor the average distance of new documents from this centroid over time. Let $\mathcal{D}_j^{(t)}$ be the set of new documents assigned to class $j$ in time window $t$. The drift metric $S_j(t)$ is the average semantic distance:
        \begin{equation}
            S_j(t) = \frac{1}{|\mathcal{D}_j^{(t)}|} \sum_{d' \in \mathcal{D}_j^{(t)}} \text{distance}(\mathbf{e}_{d'}, \mathbf{c}_j)
        \end{equation}
        A sustained increase in $S_j(t)$ the average semantic distance from their assigned class centroid indicates that the class is becoming diffuse and its definition is eroding.

        \item \textbf{Novelty Detection:} Drift may also manifest as the emergence of entirely new topics not covered by the existing classification schema. The classifier will be forced to assign these novel documents to the "least wrong" existing class, degrading system performance. To detect this, the model can periodically run the unconstrained topic modeling process (from Phase 2) on a recent sample of the data. The update model then compare the resulting emergent topics against pre-established class definitions. If a new, coherent topic emerges that has a low semantic similarity to all existing class descriptions, it signifies a "conceptual gap" in the schema and a clear instance of drift.
    \end{enumerate}
    
    \item \textbf{Tracking Performance Degradation:} The ultimate measure is a direct drop in accuracy. By continuously evaluating the classifier against a curated golden set of labeled data, any statistically significant negative trend in F1-score or accuracy serves as a definitive, albeit lagging, indicator that the model is no longer aligned with ground truth and requires immediate human intervention.
\end{itemize}

An alert from any of these systems triggers a root-cause analysis. The nature of the alert guides the response: a distributional shift may require re-calibrating few-shot examples, while a drop in semantic cohesion or the detection of a novel topic signals that the prompt's core definitions must be re-evaluated, prompting a return to the iterative refinement loop of our framework.
\end{itemize}
As the data landscape evolves, the classifier must adapt. This includes staying ahead of new adversarial techniques and periodically re-running the iterative refinement process to ensure the classification schema remains representative and relevant.

\section{Conclusion}
We have presented a comprehensive, semi-supervised framework for developing, validating, and monitoring hierarchical text classifiers using Large Language Models. Using an innovative approach to combining the structured guidance of domain knowledge with the semantic understanding of LLMs, our iterative methodology enables the creation of robust and interpretable classification systems. The emphasis on multi-faceted validation—spanning quantitative accuracy, sequence invariance, and adversarial robustness—provides a pathway for deploying these powerful models in mission-critical applications.

Beyond its methodological contributions, the practical value of this framework is realized when its outputs are translated into actionable product intelligence. By applying this classifier to streams of user feedback—such as support tickets, forum posts, or survey responses—product teams can move from anecdotal evidence to quantitative understanding. For instance:

\begin{itemize}
    \item \textbf{Proactive Bug Triage and Prioritization:} A classifier trained on user bug reports might identify a hierarchy such as \texttt{Technical Issue} $\rightarrow$ \texttt{Login Failure} $\rightarrow$ \texttt{SSO Error}. A sudden spike in the volume of the \texttt{SSO Error} class on a monitoring dashboard provides an immediate, precise signal to the engineering team. It allows them to bypass generic "login is broken" reports and focus directly on the specific failing integration, dramatically reducing triage time and mean time to resolution.

    \item \textbf{Identifying User Experience Friction:} By analyzing feedback from new users, the system could surface a consistently high volume for the class \texttt{User Confusion} $\rightarrow$ \texttt{Feature Discovery} $\rightarrow$ \texttt{Cannot Find Button}. This provides quantitative evidence that a key feature, while functional, is poorly positioned in the user interface. This insight allows the UX design team to prioritize a UI redesign and subsequently measure its success by observing if the volume of this specific complaint decreases.

    \item \textbf{Strategic Roadmap Planning:} When applied to churn surveys or feedback mentioning competitors, the classifier might reveal a growing trend in the class \texttt{Feature Gap} $\rightarrow$ \texttt{Missing Integration (Competitor X)}. This transforms vague notions of competitive weakness into a concrete, data-backed business case for prioritizing the development of a specific integration in the next product roadmap cycle.

    \item \textbf{Gauging Policy and Communication Impact:} After announcing a new platform policy, an influx of developer feedback can be classified. Discovering that the dominant class is not \texttt{Policy Disagreement} $\rightarrow$ \texttt{Policy Unfair}, but rather \texttt{Policy Disagreement} $\rightarrow$ \texttt{Communication Unclear}, provides a crucial insight. It indicates the problem is not with the policy's substance but with the clarity of its announcement. The communications team can then use the raw text from this category to write a more effective FAQ and address the specific points of confusion.
\end{itemize}

Ultimately, this framework serves as a practical guide for researchers and practitioners seeking to build more than just a classifier. It enables the creation of a dynamic sensing system that listens to users at scale, quantifies their problems and preferences, and empowers teams to build better products through data-driven, proactive decision-making to understand the product and interaction points with users.

\newpage
\bibliographystyle{alpha}
\bibliography{main}

\newcommand{\etalchar}[1]{$^{#1}$}
\begin{thebibliography}{YLMMRS21}

\bibitem[BKK{\etalchar{+}}22]{bai2022constitutional}
Yuntao Bai, Saurav Kadavath, Sandipan Kundu, Amanda Askell, Jackson Kernion, Andy Jones, Anna Chen, Anna Goldie, Azalia Mirhoseini, Cameron McKinnon, Carol Chen, Catherine Olsson, Christopher Olah, Danny Hernandez, Dawn Drain, Deep Ganguli, Dustin Li, Eli Tran-Johnson, Ethan Perez, Jamie Kerr, Jared Mueller, Jeffrey Ladish, Joshua Landau, Kamal Ndousse, Kamile Lukosuite, Liane Lovitt, Michael Sellitto, Nelson Elhage, Nicholas Schiefer, Noemi Mercado, Nova DasSarma, Robert Lasenby, Robin Larson, Sam Ringer, Scott Johnston, Shauna Kravec, Sheer~El Showk, Stanislav Fort, Tamera Lanham, Timothy Telleen-Lawton, Tom Conerly, Tom Henighan, Tristan Hume, Samuel~R. Bowman, Zac Hatfield-Dodds, Ben Mann, Dario Amodei, Nicholas Joseph, Sam McCandlish, Tom Brown, and Jared Kaplan.
\newblock Constitutional ai: Harmlessness from ai feedback, 2022.

\bibitem[CXW{\etalchar{+}}24]{chang2024efficientpromptingmethodslarge}
Kaiyan Chang, Songcheng Xu, Chenglong Wang, Yingfeng Luo, Xiaoqian Liu, Tong Xiao, and Jingbo Zhu.
\newblock Efficient prompting methods for large language models: A survey, 2024.

\bibitem[HBD{\etalchar{+}}19]{holtzman2019curious}
Ari Holtzman, Jan Buys, Li~Du, Maxwell Forbes, and Yejin Choi.
\newblock The curious case of neural text degeneration, 2019.

\bibitem[LKB{\etalchar{+}}23]{lightman2023letsverify}
Hunter Lightman, Vineet Kosaraju, Yura Burda, Harri Edwards, Bowen Baker, Teddy Lee, Jan Leike, John Schulman, Ilya Sutskever, and Karl Cobbe.
\newblock Let's verify step by step, 2023.

\bibitem[LLH{\etalchar{+}}23]{liu2023lost}
Nelson~F. Liu, Kevin Lin, John Hewitt, Ashwin Paranjape, Michele Bevilacqua, Fabio Petroni, and Percy Liang.
\newblock Lost in the middle: How language models use long contexts, 2023.

\bibitem[MTG{\etalchar{+}}23]{paskun2023selfrefine}
Aman Madaan, Niket Tandon, Prakhar Gupta, Skyler Hallinan, Luyu Gao, Sarah Wiegreffe, Uri Alon, Nouha Dziri, Shrimai Prabhumoye, Yiming Yang, Shashank Gupta, Bodhisattwa~Prasad Majumder, Katherine Hermann, Sean Welleck, Amir Yazdanbakhsh, and Peter Clark.
\newblock Self-refine: Iterative refinement with self-feedback, 2023.

\bibitem[OWJ{\etalchar{+}}22]{ouyang2022training}
Long Ouyang, Jeff Wu, Xu~Jiang, Diogo Almeida, Carroll~L. Wainwright, Pamela Mishkin, Chong Zhang, Sandhini Agarwal, Katarina Slama, Alex Ray, John Schulman, Jacob Hilton, Fraser Kelton, Luke Miller, Maddie Simens, Amanda Askell, Peter Welinder, Paul Christiano, Jan Leike, and Ryan Lowe.
\newblock Training language models to follow instructions with human feedback, 2022.

\bibitem[RSM{\etalchar{+}}23]{rafailov2023direct}
Rafael Rafailov, Archit Sharma, Eric Mitchell, Stefano Ermon, Christopher~D. Manning, and Chelsea Finn.
\newblock Direct preference optimization: Your language model is secretly a reward model, 2023.

\bibitem[VSP{\etalchar{+}}17]{vaswani2017attention}
Ashish Vaswani, Noam Shazeer, Niki Parmar, Jakob Uszkoreit, Llion Jones, Aidan~N. Gomez, Łukasz Kaiser, and Illia Polosukhin.
\newblock Attention is all you need, 2017.

\bibitem[WWS{\etalchar{+}}22]{wei2022chainofthought}
Jason Wei, Xuezhi Wang, Dale Schuurmans, Maarten Bosma, Fei Xia, Ed~Chi, Quoc~V. Le, and Denny Zhou.
\newblock Chain-of-thought prompting elicits reasoning in large language models, 2022.

\bibitem[YLMMRS21]{lu2021fantastically}
Bartolo Yao Lu~Max, Alastair Moore, Sebastian Riedel, and Pontus Stenetorp.
\newblock Fantastically ordered prompts and where to find them: Overcoming few-shot prompt order sensitivity, 2021.

\bibitem[ZWF{\etalchar{+}}21]{zhao2021calibrate}
Tony~Z. Zhao, Eric Wallace, Shi Feng, Dan Klein, and Sameer Singh.
\newblock Calibrate before use: Improving few-shot performance of language models, 2021.

\end{thebibliography}

\end{document}